# Drawing and Recognizing Chinese Characters with Recurrent Neural Network

Xu-Yao Zhang, Fei Yin, Yan-Ming Zhang, Cheng-Lin Liu, Yoshua Bengio

*Abstract*—Recent deep learning based approaches have achieved great success on handwriting recognition. Chinese characters are among the most widely adopted writing systems in the world. Previous research has mainly focused on recognizing handwritten Chinese characters. However, recognition is only one aspect for understanding a language, another challenging and interesting task is to teach a machine to automatically write (pictographic) Chinese characters. In this paper, we propose a framework by using the recurrent neural network (RNN) as both a discriminative model for recognizing Chinese characters and a generative model for drawing (generating) Chinese characters. To recognize Chinese characters, previous methods usually adopt the convolutional neural network (CNN) models which require transforming the online handwriting trajectory into image-like representations. Instead, our RNN based approach is an end-to-end system which directly deals with the sequential structure and does not require any domain-specific knowledge. With the RNN system (combining an LSTM and GRU), state-of-the-art performance can be achieved on the ICDAR-2013 competition database. Furthermore, under the RNN framework, a conditional generative model with character embedding is proposed for automatically drawing recognizable Chinese characters. The generated characters (in vector format) are human-readable and also can be recognized by the discriminative RNN model with high accuracy. Experimental results verify the effectiveness of using RNNs as both generative and discriminative models for the tasks of drawing and recognizing Chinese characters.

## I. Introduction

Reading and writing are among the most important and fundamental skills of human beings. Automatic recognition (or reading) of handwritten characters has been studied for a long time [1] and obtained great achievements during the past decades [2], [3]. However, the automatic drawing (or writing) of characters has not been studied as much, until the recent advances based on recurrent neural network for generating sequences [4]. In the development of human intelligence, the skills of reading and writing are mutual complementary. Therefore, for the purpose of machine intelligence, it would be interesting to handle them in unified framework.

Chinese characters constitute the oldest continuously used system of writing in the world. Moreover, Chinese characters have been widely used (modified or extended) in many Asian countries such as China, Japan, Korea, and so on. There are more than tens of thousands of different Chinese characters. Most of them can be well recognized by most people, however, nowadays, it is becoming more and more difficult for people to write them correctly, due to the overuse of keyboard or touch-screen based input methods. Compared with reading, writing of Chinese characters is gradually becoming a forgotten or missing skill.

For the task of automatic recognition of handwritten Chinese characters, there are two main categories of approaches: online and offline methods. With the success of deep learning [5], [6], the convolutional neural network (CNN) [7] has been widely applied for handwriting recognition. The strong priori knowledge of convolution makes the CNN a powerful tool for image classification. Since the offline characters are naturally represented as scanned images, it is natural and works well to apply CNNs to the task of offline recognition [8], [9], [10], [11]. However, in order to apply CNNs to online characters, the online handwriting trajectory should firstly be transformed to some image-like representations, such as the AMAP [12], the path signature maps [13] or the directional feature maps [14].

During the data acquisition of online handwriting, the pen-tip movements (xy-coordinates) and pen states (down or up) are automatically stored as (variable-length) sequential data. Instead of transforming them into image-like representations, we choose to deal with the raw sequential data in order to exploit the richer information it carries. In this paper, different from the traditional approaches based on CNNs, we propose to use recurrent neural networks (RNN) combined with bidirectional long short term memory (LSTM) [15], [16] and gated recurrent unit (GRU) [17] for online handwritten Chinese character recognition. RNN is shown to be very effective for English handwriting recognition [18]. For Chinese character recognition, compared with the CNN-based approaches, our method is fully end-to-end and does not require any domain-specific knowledge. State-of-the-art performance has been achieved by our method on the ICDAR-2013 competition database [19]. To the best of our knowledge, this is the first work on using RNNs for end-to-end online handwritten Chinese character recognition.

Besides the recognition (reading) task, this paper also considers the automatic drawing of Chinese characters (writing task). Under the recurrent neural network framework, a conditional generative model is used to model the distribution of Chinese handwriting, allowing the model to generate new handwritten characters by sampling from the probability distribution associated with the RNN. The study of generative models is an important and active research topic in the deep learning field [6]. Many useful generative models have been

Xu-Yao Zhang, Fei Yin, Yan-Ming Zhang and Cheng-Lin Liu are with the NLPR at Institute of Automation, Chinese Academy of Sciences, P.R. China. Email: {xyz, fyin, ymzhang, liucl}@nlpr.ia.ac.cn.

Yoshua Bengio is with the MILA lab at University of Montreal, Canada. Email: yoshua.bengio@umontreal.ca.

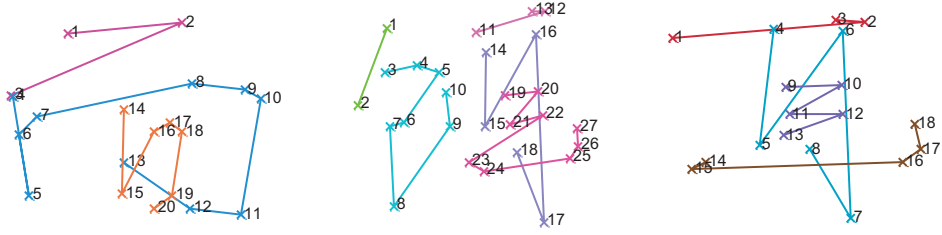

Fig. 1. Illustration of three online handwritten Chinese characters. Each color represents a stroke and the numbers indicate the writing order. The purpose of this paper is to automatically recognize and draw (generate) real and cursive Chinese characters under a single framework based on recurrent neural networks.

proposed such as NADE [20], variational auto-encoder [21], DRAW [22], and so on. To better model the generating process, the generative adversarial network (GAN) [23] simultaneously train a generator to capture the data distribution and a discriminator to distinguish real and generated samples in a min-max optimization framework. Under this framework, high-quality images can be generated with the LAPGAN [24] and DCGAN [25] models, which are extensions of the original GAN.

Recently, it was shown by [26] that realistically-looking Chinese characters can be generated with DCGAN. However, the generated characters are offline images which ignore the handwriting dynamics (temporal order and trajectory). To automatically generate the online (dynamic) handwriting trajectory, the recurrent neural network (RNN) with LSTM was shown to be very effective for English online handwriting generation [4]. The contribution of this paper is to study how to extend and adapt this technique for Chinese character generation, considering the difference between English and Chinese handwriting habits and the large number of categories for Chinese characters. As shown by [27], fake and regular-written Chinese characters can be generated under the LSTM-RNN framework. However, a more interesting and challenging problem is the generating of real (readable) and cursive handwritten Chinese characters.

To reach this goal, we propose a conditional RNN-based generative model (equipped with GRUs or LSTMs) to automatically draw human-readable cursive Chinese characters. The character embedding is jointly trained with the generative model. Therefore, given a character class, different samples (belonging to the given class but with different writing styles) can be automatically generated by the RNN model conditioned on the embedding. In this paper, the tasks of automatically drawing and recognizing Chinese characters are completed both with RNNs, seen as either generative or discriminative models. Therefore, to verify the quality of the generated characters, we can feed them into the pre-trained discriminative RNN model to see whether they can be correctly classified or not. It is found that most of the generated characters can be automatically recognized with high accuracy. This verifies the effectiveness of the proposed method in generating real and cursive Chinese characters.

The rest of this paper is organized as follows. Section II introduces the representation of online handwritten Chinese characters. Section III describes the discriminative RNN model for end-to-end recognition of handwritten Chinese characters. Section IV reports the experimental results on the ICDAR-2013 competition database. Section V details the generative RNN model for drawing recognizable Chinese characters. Section VI shows the examples and analyses of the generated characters. At last, Section VII draws the concluding remarks.

## II. REPRESENTATION FOR ONLINE HANDWRITTEN CHINESE CHARACTER

Different from the static image based representation for offline handwritten characters, rich dynamic (spatial and temporal) information can be collected in the writing process for online handwritten characters, which can be represented as a variable length sequence:

$$[[x_1, y_1, s_1], [x_2, y_2, s_2], \ldots, [x_n, y_n, s_n]], \quad (1)$$

where $x_i$ and $y_i$ are the xy-coordinates of the pen movements and $s_i$ indicates which stroke the point $i$ belongs to. As shown in Fig. 1, Chinese characters usually contain multiple strokes and each stroke is produced by numerous points. Besides the character shape information, the writing order is also preserved in the online sequential data, which is valuable and very hard to recover from the static image. Therefore, to capture the dynamic information for increasing recognition accuracy and also to improve the naturalness of the generated characters, we directly make use of the raw sequential data rather than transforming them into an image-like representation.

### A. Removing Redundant Points

Different people may have different handwriting habits (e.g., regular, fluent, cursive, and so on), resulting in significantly different number of sampling points, even when they are writing the same character. To remove the redundant points, we propose a simple strategy to preprocess the sequence. Consider a particular point $(x_i, y_i, s_i)$. Let's assume $s_i = s_{i-1} = s_{i+1}$, otherwise, it will be the starting or ending point of a stroke which will always be preserved. Whether to remove point $i$ or not depends on two conditions. The first condition is based on the distance of this point away from its former point:

$$\sqrt{(x_i - x_{i-1})^2 + (y_i - y_{i-1})^2} < T_{\text{dist}}. \quad (2)$$

As shown in Fig. 2(a), if point $i$ is too close to point $i-1$, it should be removed. Moreover, point $i$ should also be removed if it is on a straight line connecting points $i-1$ and $i+1$. Let $\triangle x_i = x_{i+1} - x_i$ and $\triangle y_i = y_{i+1} - y_i$, the second condition is based on the cosine similarity:

$$\frac{\triangle x_{i-1}\triangle x_i + \triangle y_{i-1}\triangle y_i}{(\triangle x_{i-1}^2 + \triangle y_{i-1}^2)^{0.5}(\triangle x_i^2 + \triangle y_i^2)^{0.5}} > T_{\cos}. \quad (3)$$

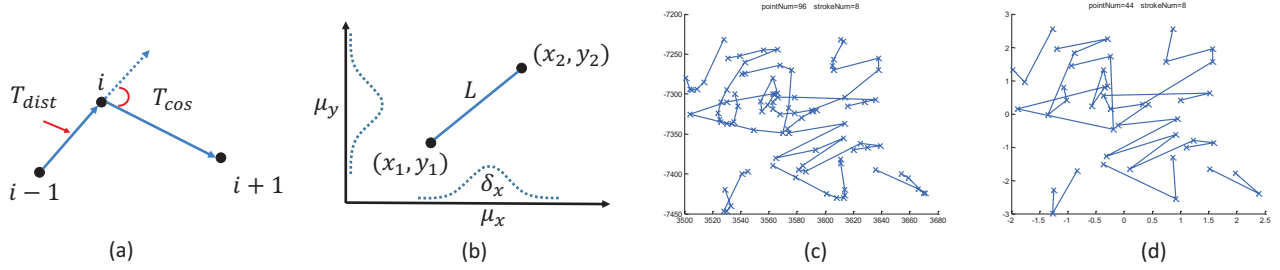

Fig. 2. (a) Removing of redundant points. (b) Coordinate normalization. (c) Character before preprocessing. (d) Character after preprocessing.

If one of the conditions in Eq. (2) or Eq. (3) is satisfied, the point $i$ should be removed. With this preprocessing, the shape information of the character is still well preserved, but each item (point) in the new sequence becomes more informative, since the redundant points have been removed.

### B. Coordinate Normalization

Another influence is coming from the variations in size or absolute values of the coordinates for the characters captured with different devices or written by different people. Therefore, we must normalize the xy-coordinates into a standard interval. Specifically, as shown in Fig. 2(b), consider a straight line $L$ connecting two points $(x_1, y_1)$ and $(x_2, y_2)$, the projections of this line onto x-axis and y-axis are

$$\begin{aligned} p_x(L) &= \int_L x dL = \frac{1}{2}\text{len}(L)(x_1 + x_2), \\ p_y(L) &= \int_L y dL = \frac{1}{2}\text{len}(L)(y_1 + y_2), \end{aligned} \quad (4)$$

where $\text{len}(L) = \sqrt{(x_2-x_1)^2 + (y_2-y_1)^2}$ denotes the length of $L$. With these information, we can estimate the mean values by projecting all lines onto x-axis and y-axis:

$$\mu_x = \frac{\sum_{L\in\Omega} p_x(L)}{\sum_{L\in\Omega} \text{len}(L)}, \quad \mu_y = \frac{\sum_{L\in\Omega} p_y(L)}{\sum_{L\in\Omega} \text{len}(L)}, \quad (5)$$

where $\Omega$ represents the set of all straight lines that connect two successive points within the same stroke. After this, we estimate the deviation (from mean) of the projections:

$$\begin{aligned} d_x(L) = \int_L (x-\mu_x)^2 dL = \frac{1}{3}\text{len}(L)\bigl[(x_2-\mu_x)^2 + \\ (x_1-\mu_x)^2 + (x_1-\mu_x)(x_2-\mu_x)\bigr]. \end{aligned} \quad (6)$$

The standard deviation on x-axis can then be estimated as:

$$\delta_x = \sqrt{\frac{\sum_{L\in\Omega} d_x(L)}{\sum_{L\in\Omega} \text{len}(L)}}. \quad (7)$$

With all the information of $\mu_x, \mu_y$ and $\delta_x$ estimated from one character, we can now normalize the coordinates by:

$$x_{\text{new}} = (x-\mu_x)/\delta_x, \quad y_{\text{new}} = (y-\mu_y)/\delta_x. \quad (8)$$

This normalization is applied globally for all the points in the character. Note that we do not estimate the standard deviation on the y-axis and the y-coordinate is also normalized by $\delta_x$. The reason for doing so is to keep the original ratio of height and width for the character, and also keep the writing direction for each stroke. After coordinate normalization, each character is placed in a standard xy-coordinate system, while the shape of the character is kept unchanged.

### C. Illustration

The characters before and after preprocessing are illustrated in Fig. 2 (c) and (d) respectively. It is shown that the character shape is well preserved, and many redundant points have been removed. The original character contains 96 points while the processed character only has 44 points. This will make each point more informative, which will benefit RNN modeling not just because of speed but also because the issue of long-term dependencies [28] is thus reduced because sequences are shorter. Moreover, as shown in Fig. 2(d), the coordinates of the new character is normalized. In the new coordinate system, the position of $(0,0)$ is located in the central part of the character, and the deviations on the xy-axis are also normalized. Since in this paper we use a sequence-based representation, the preprocessing used here is different from the traditional methods designed for image-based representation, such as the equidistance sampling [3] and character shape normalization [29].

## III. DISCRIMINATIVE MODEL: END-TO-END RECOGNITION WITH RECURRENT NEURAL NETWORK

The best established approaches for recognizing Chinese characters are based on transforming the sequence of Eq. (1) into some image-like representation [12], [13], [14] and then applying the convolutional neural network (CNN). For the purpose of fully end-to-end recognition, we apply a recurrent neural network (RNN) directly on the raw sequential data. Due to better utilization of temporal and spatial information, our RNN approach can achieve higher accuracy than previous CNN and image-based models.

### A. Representation for Recognition

From the sequence of Eq. (1) (after preprocessing), we extract a six-dimensional representation for each straight line $L_i$ connecting two points $i$ and $i+1$:

$$L_i = [x_i, y_i, \triangle x_i, \triangle y_i, \mathbb{I}(s_i = s_{i+1}), \mathbb{I}(s_i \neq s_{i+1})], \quad (9)$$

where $\triangle x_i = x_{i+1} - x_i$, $\triangle y_i = y_{i+1} - y_i$ and $\mathbb{I}(\cdot) = 1$ when the condition is true and otherwise zero. In each $L_i$, the first two terms are the start position of the line, and the 3-4th terms are the direction of the pen movements, while the last two terms indicate the status of the pen, i.e., $[1, 0]$ means pen-down while $[0, 1]$ means pen-up. With this representation, the character in Eq. (1) is transformed to a new sequence of $[L_1, L_2, \ldots, L_{n-1}]$. To simplify the notations used in following subsections, we will use $[x_1, x_2, \ldots, x_k]$ to denote a general sequence, but note that each item $x_i$ here is actually the six-dimensional vector shown in Eq. (9).

### B. Recurrent Neural Network (RNN)

The RNN is a natural generalization of the feedforward neural networks to sequences [30]. Given a general input sequence $[x_1, x_2, \ldots, x_k]$ where $x_i \in \mathbb{R}^d$ (different samples may have different sequence length $k$), at each time-step of RNN modeling, a hidden state is produced, resulting in a hidden sequence of $[h_1, h_2, \ldots, h_k]$. The activation of the hidden state at time-step $t$ is computed as a function $f$ of the current input $x_t$ and previous hidden state $h_{t-1}$ as:

$$h_t = f(x_t, h_{t-1}). \qquad (10)$$

At each time-step, an optional output can be produced by $y_t = g(h_t)$, resulting in an output sequence $[y_1, y_2, \ldots, y_k]$, which can be used for sequence-to-sequence tasks, for example, based on the CTC framework [31]. In this section, the input sequence is encoded into a fixed-length vector for final classification, due to the recursively applied transition function $f$. The RNN computes activations for each time-step which makes them extremely deep and can lead to vanishing or exploding gradients [28]. The choice of the recurrent computation $f$ can have a big impact on the success of RNN because the spectrum of its Jacobian controls whether gradients tend to propagate well (or vanish or explode). In this paper, we use both long short term memory (LSTM) [15] [16] and gated recurrent unit (GRU) [17] for RNN modeling.

### C. Long Short Term Memory (LSTM)

LSTM [15] [16] is widely applied because it reduces the vanishing and exploding gradient problems and can learn longer term dependencies. With LSTMs, for time-step $t$, there is an input gate $i_t$, forget gate $f_t$, and output gate $o_t$:

$$i_t = \text{sigm}(W_i x_t + U_i h_{t-1} + b_i), \qquad (11)$$
$$f_t = \text{sigm}(W_f x_t + U_f h_{t-1} + b_f), \qquad (12)$$
$$o_t = \text{sigm}(W_o x_t + U_o h_{t-1} + b_o), \qquad (13)$$
$$\widetilde{c}_t = \tanh(W_c x_t + U_c h_{t-1} + b_c), \qquad (14)$$
$$c_t = i_t \odot \widetilde{c}_t + f_t \odot c_{t-1}, \qquad (15)$$
$$h_t = o_t \odot \tanh(c_t), \qquad (16)$$

where $W_*$ is the input-to-hidden weight matrix, $U_*$ is the state-to-state recurrent weight matrix, and $b_*$ is the bias vector. The operation $\odot$ denotes the element-wise vector product. The hidden state of LSTM is the concatenation of $(c_t, h_t)$. The long-term memory is saved in $c_t$, and the forget gate and input gate are used to control the updating of $c_t$ as shown in Eq. (15), while the output gate is used to control the updating of $h_t$ as shown in Eq. (16).

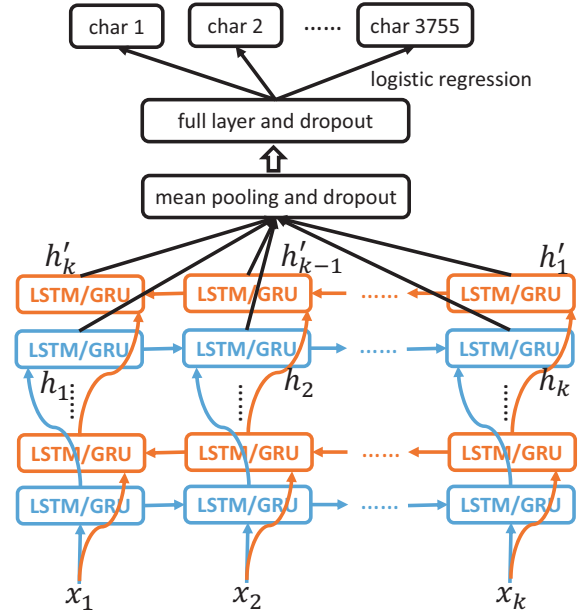

Fig. 3. The stacked bidirectional RNN for end-to-end recognition.

### D. Gated Recurrent Unit (GRU)

RNNs with gated recurrent units (GRU) [17] can be viewed as a light-weight version of LSTMs. Similar to the LSTM unit, the GRU also has gating units (reset gate $r_t$ and update gate $z_t$) that modulate the flow of information inside the unit, however, without having a separate memory cell.

$$r_t = \text{sigm}(W_r x_t + U_r h_{t-1} + b_r), \qquad (17)$$
$$z_t = \text{sigm}(W_z x_t + U_z h_{t-1} + b_z), \qquad (18)$$
$$\widetilde{h}_t = \tanh(W x_t + U(r_t \odot h_{t-1}) + b), \qquad (19)$$
$$h_t = z_t \odot h_{t-1} + (1 - z_t) \odot \widetilde{h}_t. \qquad (20)$$

The activation of GRU $h_t$ is a linear interpolation between the previous activation $h_{t-1}$ and the candidate activation $\widetilde{h}_t$, controlled by the update gate $z_t$. As shown in Eq. (19), when reset gate $r_t$ is off (close to zero), the GRU acts like reading the first symbol of an input sequence, allowing it to forget the previously computed state. It has been shown that GRUs and LSTMs have similar performance [32].

### E. Stacked and Bidirectional RNN

In real applications, contexts from both past and future are useful and complementary to each other [18]. Therefore, we combine forward (left to right) and backward (right to left) recurrent layers to build a bidirectional RNN model [33]. Moreover, the stacked recurrent layers are used to build a deep RNN system. As shown in Fig. 3, by passing $[x_1, x_2, \ldots, x_k]$ through the forward recurrent layers, we can obtain a hidden state sequence of $[h_1, h_2, \ldots, h_k]$. Meanwhile, by passing the reversed sequence of $[x_k, x_{k-1}, \ldots, x_1]$ through the backward recurrent layers, we can get another hidden state sequence of $[h'_1, h'_2, \ldots, h'_k]$. To make a final classification, all the hidden states are combined to obtain a fixed-length representation for the input sequence:

$$\text{Fixed Length Feature} = \frac{1}{2k} \sum_{i=1}^{k} (h_i + h'_i), \quad (21)$$

which is then fed into a fully connected layer and a softmax layer for final classification. The whole model can be efficiently and effectively trained by minimizing the multi-class negative log-likelihood loss with stochastic gradient descent, using the back-propagation algorithm [34] to compute gradients.

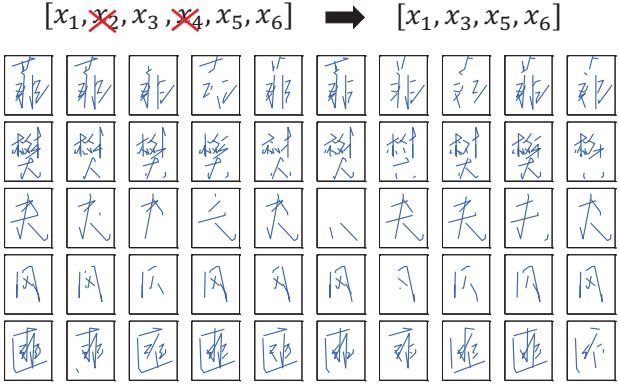

Fig. 4. Illustration of data augmentation by sequential dropout on the input sequence. The first column shows the original character, and the remaining columns are the characters after random dropout with probability 0.3.

### F. Regularization and Data Augmentation

Regularization is important for improving the generalization performance of deep neural network. As shown in Fig. 3, we apply the dropout [35] strategy on both the mean-pooling layer and the fully connected layer. Another key to the success of deep neural network is the large number of training data. Recently, it is shown that the dropout can also be viewed as some kind of data augmentation [36]. For traditional image-based recognition system, random distortion is widely used as the data augmentation strategy [13], [14]. In this paper, we use a simple strategy to regularize and augment data for sequence classification, which we call **sequential dropout**. As shown in Fig. 4, given a sequence, many sub-sequences can be generated by randomly removing some items in the original sequence with a given probability. This of course could make more sense for some distributions and worked well for our data. For this to work, the preserved sub-sequence must still contain enough information for categorization, as shown in Fig. 4. This strategy is similar to the previous proposed dropStroke [37] and dropSegment [38] methods in handwriting analysis. However, our approach is much simpler and general, not requiring any domain-specific knowledge (e.g. stroke/segment detection) in order to identify pieces to be dropped out.

With sequential dropout on the input, we can build a large enough (or infinite) training set, where each training sequence is only shown once. In the testing process, two strategies can be used. First, we can directly feed the full-sequence into the RNN for classification. Second, we can also apply sequential dropout to obtain multiple sub-sequences, and then make an ensemble-based decision by fusing the classification results from these sub-sequences. The comparison of these two approaches will be discussed in the experimental section.

### G. Initialization and Optimization

Initialization is very important for deep neural networks. We initialize all the weight matrices in LSTM/GRU ($W_*$ and $U_*$), full layer, and logistic regression layer by random values drawn from the zero-mean Gaussian distribution with standard deviation 0.01. All bias terms are initialized as zeros, except the forget gate in LSTM. As suggested by [39], we initialize the forget gate bias $b_f$ to be a large value of 5. The purpose of doing so is to make sure that the forget gate in Eq. (12) is initialized close to one (which means no forgetting), and then long-range dependencies can be better learned in the beginning of training. The cell and hidden states of LSTMs and GRUs are initialized at zero. Optimization is another important issue for deep learning. In this paper, we use a recently proposed first-order gradient method called Adam [40] which is based on adaptive estimation of lower-order moments. These strategies helped to make the training of RNNs to be both efficient and effective.

## IV. EXPERIMENTS ON RECOGNIZING CHINESE CHARACTERS

In this section, we present experiments on recognizing cursive online handwritten Chinese characters, for the purpose of evaluating and comparing the proposed discriminative RNN model with other state-of-the-art approaches.

### A. Database

The database used for evaluation is from the ICDAR-2013 competition [19] of online Chinese handwriting recognition, which is a third version of the previous competitions held on CCPR-2010 [41] and ICDAR-2011 [42]. The database used for training is the CASIA database [43] including OLHWDB1.0 and OLHWDB1.1. There are totally 2,693,183 samples for training and 224,590 samples for testing. The training and test data were produced by different writers. The number of character class is 3,755 (level-1 set of GB2312-80). Online handwritten Chinese character recognition is a challenging problem [3] due to the large number of character classes, confusion between many similar characters, and distinct handwriting styles across individuals. Many teams from both academia and industry were involved in the three competitions, and the recognition accuracy had been promoted gradually and significantly through the competitions [41], [42], [19].

### B. Implementation Details

In this paper, each character is represented by a sequence as shown in Eq. (1). The two hyper-parameters used in Section II for preprocessing are $T_{\text{dist}} = 0.01 \times \max\{H, W\}$ and $T_{\cos} = 0.99$, where $H$ is the height and $W$ is the width of the character. After preprocessing, the average length of the sequences for each character is about 50. As shown in Fig. 3, to increase the generalization performance, the dropout is used for the mean pooling layer and full layer with probability 0.1. Moreover, dropout (with probability 0.3) is also used on input sequence for data augmentation as described in Section III-F. The initialization of the network is described in Section III-G. The optimization algorithm is the Adam [40] with mini-batch size 1000. The initial learning rate is set to be 0.001 and then decreased by ×0.3 when the cost or accuracy on the training data stop improving. After each epoch, we shuffle the training data to make different mini-batches. All the models were implemented under the Theano [44], [45] platform using the NVIDIA Titan-X 12G GPU.

TABLE I
COMPARISON OF DIFFERENT NETWORK ARCHITECTURES FOR ONLINE HANDWRITTEN CHINESE CHARACTER RECOGNITION.

| Name | Architecture | Recurrent Type | Memory | Train Time | Test Speed | Train Acc. | Test Acc. |
|------|--------------|----------------|--------|------------|------------|------------|-----------|
| NET1 | $6 \to [500] \to 200 \to 3755$ | LSTM | 11.00MB | 95.16h | 0.3792ms | 98.07% | 97.67% |
| NET2 | $6 \to [500] \to 200 \to 3755$ | GRU | 9.06MB | 75.92h | 0.2949ms | 97.81% | 97.71% |
| NET3 | $6 \to [100, 500] \to 200 \to 3755$ | LSTM | 12.76MB | 125.43h | 0.5063ms | 98.21% | 97.70% |
| NET4 | $6 \to [100, 500] \to 200 \to 3755$ | GRU | 10.38MB | 99.77h | 0.3774ms | 97.87% | 97.76% |
| NET5 | $6 \to [100, 300, 500] \to 200 \to 3755$ | LSTM | 19.48MB | 216.75h | 0.7974ms | 98.67% | 97.80% |
| NET6 | $6 \to [100, 300, 500] \to 200 \to 3755$ | GRU | 15.43MB | 168.80h | 0.6137ms | 97.97% | 97.77% |

TABLE II
TEST ACCURACIES (%) OF ENSEMBLE-BASED DECISIONS FROM SUB-SEQUENCES GENERATED BY RANDOM DROPOUT.

| Name | Full | Ensemble of Sub-Sequences | | | | | |
|------|------|---|---|---|---|---|---|
| | | 1 | 5 | 10 | 15 | 20 | 30 |
| NET1 | 97.67 | 96.53 | 97.74 | 97.82 | 97.84 | 97.84 | 97.86 |
| NET2 | 97.71 | 96.56 | 97.77 | 97.84 | 97.86 | 97.85 | 97.89 |
| NET3 | 97.70 | 96.56 | 97.71 | 97.82 | 97.84 | 97.85 | 97.86 |
| NET4 | 97.76 | 96.54 | 97.78 | 97.86 | 97.87 | 97.88 | 97.89 |
| NET5 | 97.80 | 96.79 | 97.82 | 97.91 | 97.93 | 97.94 | 97.96 |
| NET6 | 97.77 | 96.64 | 97.79 | 97.87 | 97.88 | 97.89 | 97.91 |

### C. Experimental Results

Table I shows the comparison of different network architectures which can be represented by a general form as:

$$A \to [B_1, \ldots, B_n] \to C \to D. \tag{22}$$

The symbol $A$ is the dimension for each element in the input sequence. Moreover, the $[B_1, \ldots, B_n]$ represents $n$ stacked bidirectional recurrent layers as shown in Fig. 3, and $B_i$ is the dimension for the hidden states of LSTM or GRU at the $i$-th recurrent layer. The symbol $C$ is the number of hidden units for the full layer, and $D$ is the number of units for the logistic regression layer (also the number of character classes). The 3th column in Table I shows the recurrent unit type (LSTM or GRU) for each model. Different networks are compared from five aspects including: memory consumption in 4th column, total training time (in hours) in 5th column, evaluation/testing speed (in millisecond) for one character in 6th column, training accuracy in 7th column, and test accuracy in the last column. Other configurations are totally the same as described in Section IV-B to give a fair comparison of different architectures. It is shown that the best performance (test accuracy) is achieved by NET5, while NET4 and NET6 are very competitive with NET5.

### D. Comparison of LSTM and GRU

As shown in Sections III-C and III-D, both LSTM and GRU adopt the gating strategy to control the information flow, and therefore allow the model to capture long-term dependence embedded in the sequential data. In this paper, multiple RNN models with either LSTM or GRU recurrent units were trained and compared under the same configurations. As shown in Table I, from the perspective of test accuracy, NET2 outperforms NET1, NET4 beats NET3, and NET5 is better than NET6. However, the differences are not significant. Therefore, the only conclusion we can drawn is that LSTM and GRU have comparable prediction accuracies for our classification task. Another finding is that LSTM usually leads to higher training accuracy but not necessarily higher test accuracy. This may suggest that GRU has some ability to avoid over-fitting. Furthermore, as revealed in Table I, from the perspectives of memory consumption, training time, and especially testing speed, we can conclude that GRU is much better than LSTM. The GRU can be viewed as a light-weight version of LSTM, and still shares similar functionalities with LSTM, which makes GRU favoured by practical applications with particular requirements on memory or speed.

### E. Comparison of Different Depths

As described in Section III-E, the stacked bidirectional recurrent layers are used to build the deep RNN systems. Different depths for the networks are also compared in Table I. Compared with only one bidirectional recurrent layer (NET1 and NET2), stacking two layers (NET3 and NET4) and three layers (NET5 and NET6) can indeed improve both the training and test accuracies. However, the improvements are not significant and also vanishing when more layers being stacked. This is because the recurrent units maintain activations for each time-step which already make the RNN model to be extremely deep, and therefore, stacking more layers will not bring too much additional discriminative ability to the model. Moreover, as shown in Table I, with more stacked recurrent layers, both the training and testing time are increased dramatically. Therefore, we did not consider more than three stacked recurrent layers. From the perspectives of both accuracy and efficiency, in real applications, NET4 is preferred among the six different network architectures.

TABLE III
RESULTS ON ICDAR-2013 COMPETITION DATABASE OF ONLINE
HANDWRITTEN CHINESE CHARACTER RECOGNITION.

| Methods: | Ref. | Memory | Accuracy |
|---|---|---|---|
| Human Performance | [19] | n/a | 95.19% |
| Traditional Benchmark | [46] | 120.0MB | 95.31% |
| ICDAR-2011 Winner: VO-3 | [42] | 41.62MB | 95.77% |
| ICDAR-2013 Winner: UWarwick | [19] | 37.80MB | 97.39% |
| ICDAR-2013 Runner-up: VO-3 | [19] | 87.60MB | 96.87% |
| DropSample-DCNN | [14] | 15.00MB | 97.23% |
| DropSample-DCNN-Ensemble-9 | [14] | 135.0MB | 97.51% |
| RNN: NET4 | ours | 10.38MB | 97.76% |
| RNN: NET4-subseq30 | ours | 10.38MB | 97.89% |
| RNN: Ensemble-NET123456 | ours | 78.11MB | **98.15%** |

*F. Ensemble-based Decision from Sub-sequences*

In our experiments, the dropout [35] strategy is applied not only as a regularization (Fig. 3) but also as a data augmentation method (Section III-F). As shown in Fig. 4, in the testing process, we can still apply dropout on the input sequence to generate multiple sub-sequences, and then make ensemble-based decisions to further improve the accuracy. Specifically, the probabilities (outputs from one network) of each sub-sequence are averaged to make the final prediction.

Tabel II reports the results for ensemble-based decisions from sub-sequences. It is shown that with only one sub-sequence, the accuracy is inferior compared with the full-sequence. This is easy to understand, since there exists information loss in each sub-sequence. However, with more and more randomly sampled sub-sequences being involved in the ensemble, the classification accuracies are gradually improved. Finally, with the ensemble of 30 sub-sequences, the accuracies for different networks become consistently higher than the full-sequence based prediction. These results verified the effectiveness of using dropout for ensemble-based sequence classification.

*G. Comparison with Other State-of-the-art Approaches*

To compare our method with other approaches, Table III lists the state-of-the-art performance achieved by previous works on the ICDAR-2013 competition database [19]. It is shown that the deep learning based approaches outperform the traditional methods with large margins. In the ICDAR-2011 competition, the winner is the Vision Objects Ltd. from France using a multilayer perceptron (MLP) classifier. Moreover, in ICDAR-2013, the winner is from University of Warwick, UK, using the path signature feature map and a deep convolutional neural network [13]. Recently, the state-of-the-art performance has been achieved by [14] with domain-specific knowledge and the ensemble of nine convolutional neural networks.

As revealed in Table III and Table I, all of our models (from NET1 to NET6) can easily outperform previous benchmarks. Taking NET4 as an example, compared with other approaches, it is better from the aspects of both memory consumption and classification accuracy. The previous best performance has usually been achieved with convolutional neural networks (CNN), which has a particular requirement of transforming the online sequential handwriting data into some image-like representations [12], [13], [14]. On the contrary, our discriminative RNN model directly deals with the raw sequential data, and therefore has the potential to exploit additional information which is discarded in the spatial representations. Moreover, our method is also fully end-to-end, depending only on generic priors about sequential data processing, and not requiring any other domain-specific knowledge. These results suggest that: compared with CNNs, RNNs should be the first choice for online handwriting recognition, due to their powerful ability in sequence processing and the natural sequential property of online handwriting.

As shown in Table III, the ensemble of 30 sub-sequences with NET4 (NET4-subseq30) running on different random draws of sequential dropout (as discussed in Section III-F) can further improve the performance of NET4. An advantage for this ensemble is that only one trained model is required which will save the time and memory resources, in comparison with usual ensembles where multiple models should be trained. However, a drawback is that the number of randomly sampled sub-sequences should be large enough to guarantee the ensemble performance, which will be time-consuming for evaluation. Another commonly used type of ensemble is obtained by model averaging from separately trained models. The classification performance by combining the six pre-trained networks (NET1 to NET6) is shown in the last row of Table III. Due to the differences in network depths and recurrent types, the six networks are complementary to each other. Finally, with this kind of ensemble, this paper reached the accuracy of 98.15%, which is a new state-of-the-art and significantly outperforms all previously reported results for online handwritten Chinese character recognition.

## V. GENERATIVE MODEL: AUTOMATIC DRAWING RECOGNIZABLE CHINESE CHARACTERS

Given an input sequence $x$ and the corresponding character class $y$, the purpose of a discriminative model (as described in Section III) is to learn $p(y|x)$. On the other hand, the purpose of a generative model is to learn $p(x)$ or $p(x|y)$ (conditional generative model). In other words, by modeling the distribu-tion of the sequences, the generative model can be used to draw (generate) new handwritten characters automatically.

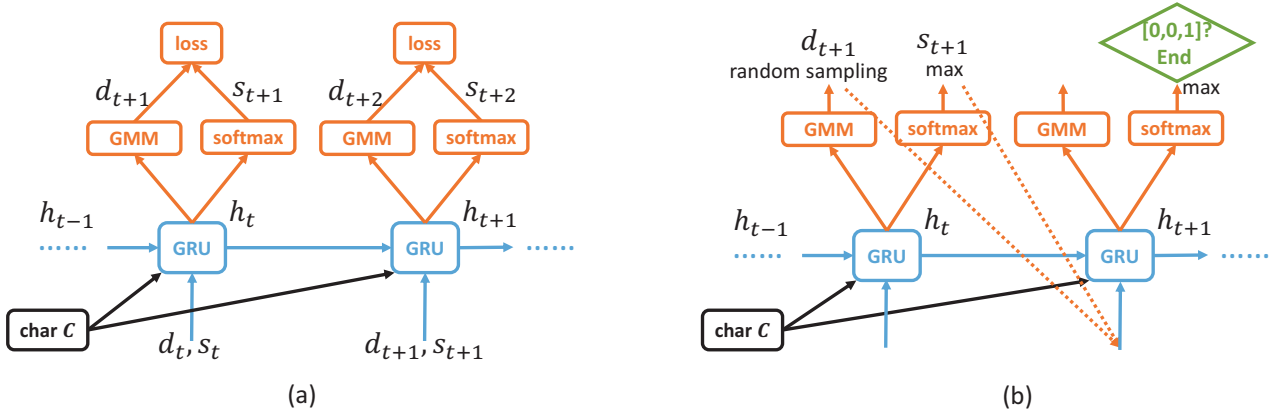

Fig. 5. For the time-step $t$ in the generative RNN model: (a) illustration of the training process, and (b) illustration of the drawing/generating process.

### A. Representation for Generation

Compared with the representation used in Section III-A, the representation for generating characters is slightly different. Motivated by [4], each character can be represented as:

$$[[d_1, s_1], [d_2, s_2], \ldots, [d_k, s_k]], \quad (23)$$

where $d_i = [\triangle x_i, \triangle y_i] \in \mathbb{R}^2$ is the pen moving direction which can be viewed as a straight line. We can draw this character with multiple lines by concatenating $[d_1, d_2, \ldots, d_k]$, i.e., the ending position of previous line is the starting position of current line. Since one character usually contains multiple strokes, each line $d_i$ may be either pen-down (should be drawn on the paper) or pen-up (should be ignored). Therefore, another variable $s_i \in \mathbb{R}^3$ is used to represent the status of pen. As suggested by [27], three states should be considered:[1]

$$s_i = \begin{cases} [1,0,0], & \text{pen-down,} \\ [0,1,0], & \text{pen-up,} \\ [0,0,1], & \text{end-of-char.} \end{cases} \quad (24)$$

With the end-of-char value of $s_i$, the RNN can automatically decide when to finish the generating process. Using the representation in Eq. (23), the character can be drawn in vector format, which is more plausible and natural than a bitmap image.

### B. Conditional Generative RNN Model

To model the distribution of the handwriting sequence, a generative RNN model is utilized. Considering that there are a large number of different Chinese characters, and in order to generate real and readable characters, the character embedding is trained jointly with the RNN model. The character embedding is a matrix $E \in \mathbb{R}^{d \times N}$ where $N$ is the number of character classes and $d$ is the embedded dimensionality. Each column in $E$ is the embedded vector for a particular class.

[1]Note that the symbol $s_i$ used here has a different meaning with the $s_i$ used in Eq. (1), and the $s_i$ used here can be easily deduced from Eq. (1).

In the following descriptions, we use $c \in \mathbb{R}^d$ to represent the embedding vector for a general character class.

Our previous experiments show that, GRUs and LSTMs have comparable performance, but the computation of GRU is more efficient. Therefore, we build our generative RNN model based on GRUs [17] rather than LSTMs. As shown in Fig. 5, at time-step $t$, the inputs for a GRU include:

- previous hidden state $h_{t-1} \in \mathbb{R}^D$,
- current pen-direction $d_t \in \mathbb{R}^2$,
- current pen-state $s_t \in \mathbb{R}^3$,
- character embedding $c \in \mathbb{R}^d$.

Following the GRU gating strategy, the updating of the hidden state and the computation of output for time-step $t$ are:

$$d'_t = \tanh\left(W_d d_t + b_d\right), \quad (25)$$
$$s'_t = \tanh\left(W_s s_t + b_s\right), \quad (26)$$
$$r_t = \text{sigm}\left(W_r h_{t-1} + U_r d'_t + V_r s'_t + M_r c + b_r\right), \quad (27)$$
$$z_t = \text{sigm}\left(W_z h_{t-1} + U_z d'_t + V_z s'_t + M_z c + b_z\right), \quad (28)$$
$$\widetilde{h}_t = \tanh\left(W(r_t \odot h_{t-1}) + U d'_t + V s'_t + M c + b\right), \quad (29)$$
$$h_t = z_t \odot h_{t-1} + (1 - z_t) \odot \widetilde{h}_t, \quad (30)$$
$$o_t = \tanh\left(W_o h_t + U_o d'_t + V_o s'_t + M_o c + b_o\right), \quad (31)$$

where $W_*, U_*, V_*, M_*$ are weight matrices and $b_*$ is the weight vector for GRU. Since both the pen-direction $d_t$ and pen-state $s_t$ are low-dimensional, we first transform them into higher-dimensional spaces by Eqs. (25) and (26). After that, the reset gate $r_t$ is computed in Eq. (27) and update gate $z_t$ is computed in Eq. (28). The candidate hidden state in Eq. (29) is controlled by the reset gate $r_t$ which can automatically decide whether to forget previous state or not. The new hidden state $h_t$ in Eq. (30) is then updated as a combination of previous and candidate hidden states controlled by update gate $z_t$. At last, an output vector $o_t$ is calculated in Eq. (31). To improve the generalization performance, the dropout strategy [35] is also applied on $o_t$.

In all these computations, the character embedding $c$ is provided to remind RNN that this is the drawing of a particular character other than random scrawling. The dynamic writing information for this character is encoded with the hidden state of the RNN, which is automatically updated and controlled by the GRU (remember or forget) due to the gating strategies. At each time-step, an output $o_t$ is produced based on new hidden state as shown in Eq. (31). From this output, the next pen-direction and pen-state should be inferred to continue the task of automatic drawing.

### C. GMM Modeling of Pen-Direction: From $o_t$ To $d_{t+1}$

As suggested by [4], the Gaussian mixture model (GMM) is used for the pen-direction. Suppose there are $M$ components in the GMM, a $5 \times M$-dimensional vector is then calculated based on the output vector $o_t$ as:

$$\left(\{\hat{\pi}^j, \hat{\mu}_x^j, \hat{\mu}_y^j, \hat{\delta}_x^j, \hat{\delta}_y^j\}_{j=1}^M\right) \in \mathbb{R}^{5M} = W_{\text{gmm}} \times o_t + b_{\text{gmm}}, \quad (32)$$

$$\pi^j = \frac{\exp(\hat{\pi}^j)}{\sum_{j'} \exp(\hat{\pi}^{j'})} \Rightarrow \pi^j \in (0,1), \sum_j \pi^j = 1, \quad (33)$$

$$\mu_x^j = \hat{\mu}_x^j \Rightarrow \mu_x^j \in \mathbb{R}, \quad (34)$$

$$\mu_y^j = \hat{\mu}_y^j \Rightarrow \mu_y^j \in \mathbb{R}, \quad (35)$$

$$\delta_x^j = \exp\left(\hat{\delta}_x^j\right) \Rightarrow \delta_x^j > 0, \quad (36)$$

$$\delta_y^j = \exp\left(\hat{\delta}_y^j\right) \Rightarrow \delta_y^j > 0. \quad (37)$$

Note the above five variables are for the time-step $t+1$, and here we omit the subscript of $t+1$ for simplification. For the $j$-th component in GMM, $\pi^j$ denotes the component weight, $\mu_x^j$ and $\mu_y^j$ denotes the means, while $\delta_x^j$ and $\delta_y^j$ are the standard deviations. The probability density $P_d(d_{t+1})$ for the next pen-direction $d_{t+1} = [\triangle x_{t+1}, \triangle y_{t+1}]$ is defined as:

$$P_d(d_{t+1}) = \sum_{j=1}^M \pi^j \mathcal{N}\left(d_{t+1}|\mu_x^j, \mu_y^j, \delta_x^j, \delta_y^j\right)$$
$$= \sum_{j=1}^M \pi^j \mathcal{N}\left(\triangle x_{t+1}|\mu_x^j, \delta_x^j\right) \mathcal{N}\left(\triangle y_{t+1}|\mu_y^j, \delta_y^j\right), \quad (38)$$

where

$$\mathcal{N}(x|\mu,\delta) = \frac{1}{\delta\sqrt{2\pi}} \exp\left(-\frac{(x-\mu)^2}{2\delta^2}\right). \quad (39)$$

Differently from [4], here for each mixture component, the x-axis and y-axis are assumed to be independent, which will simplify the model and still gives similar performance compared with the full bivariate Gaussian model. Using a GMM for modeling the pen-direction can capture the dynamic information of different handwriting styles, and hence allow RNNs to generate diverse handwritten characters.

### D. SoftMax Modeling of Pen-State: From $o_t$ To $s_{t+1}$

To model the discrete pen-states (pen-down, pen-up, or end-of-char), the softmax activation is applied on the transformation of $o_t$ to give a probability for each state:

$$\left(\hat{p}_{t+1}^1, \hat{p}_{t+1}^2, \hat{p}_{t+1}^3\right) \in \mathbb{R}^3 = W_{\text{softmax}} \times o_t + b_{\text{softmax}}, \quad (40)$$

$$p_{t+1}^i = \frac{\exp(\hat{p}_{t+1}^i)}{\sum_{j=1}^3 \exp(\hat{p}_{t+1}^j)} \in (0,1) \Rightarrow \sum_{i=1}^3 p_{t+1}^i = 1. \quad (41)$$

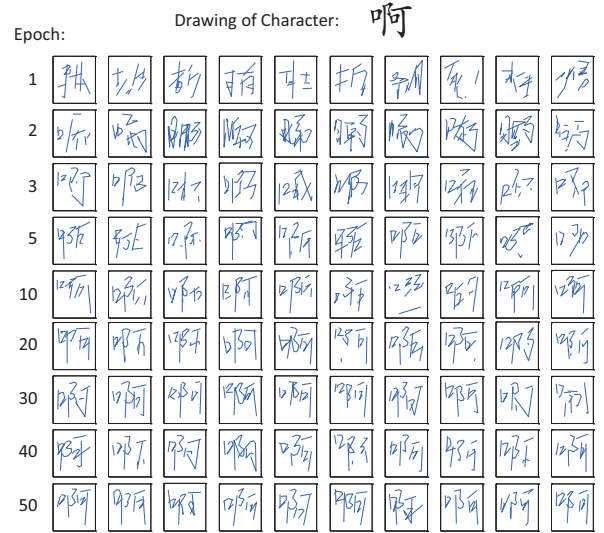

Fig. 6. Illustration of the generating/drawing for one particular character in different epochs of the training process.

The probability density $P_s(s_{t+1})$ for the next pen-state $s_{t+1} = [s_{t+1}^1, s_{t+1}^2, s_{t+1}^3]$ is then defined as:

$$P_s(s_{t+1}) = \sum_{i=1}^3 s_{t+1}^i p_{t+1}^i. \quad (42)$$

With this softmax modeling, RNN can automatically decide the status of pen and also the ending time of the generating process, according to the dynamic changes in the hidden state of GRU during drawing/writing process.

### E. Training of the Generative RNN Model

To train the generative RNN model, a loss function should be defined. Given a character represented by a sequence in Eq. (23) and its corresponding character embedding $c \in \mathbb{R}^d$, by passing them through the RNN model, as shown in Fig. 5(a), the final loss can be defined as the summation of the losses at each time-step:

$$\text{loss} = -\sum_t \left\{\log\left(P_d(d_{t+1})\right) + \log\left(P_s(s_{t+1})\right)\right\}. \quad (43)$$

However, as shown by [27], directly minimizing this loss function will lead to poor performance, because the three pen-states in Eq. (24) are not equally happened in the training process. The occurrence of pen-down is too frequent which always dominate the loss, especially compared with "end-of-char" state which only occur once for each character. To reduce the influence from this unbalanced problem, a cost-sensitive approach [27] should be used to define a new loss:

$$\text{loss} = -\sum_t \left\{\log\left(P_d(d_{t+1})\right) + \sum_{i=1}^3 w^i s_{t+1}^i \log(p_{t+1}^i)\right\}, \quad (44)$$

where $[w^1, w^2, w^3] = [1, 5, 100]$ are the weights for the losses of pen-down, pen-up, and end-of-char respectively. In this way, the RNN can be trained effectively to produce real characters. Other strategies such as initialization and optimization are the same as Section III-G.

## F. Automatic Drawing of Recognizable Characters

After training, the model can be used for automatic drawing of handwritten characters. Since this is a conditional generative model, we can first select which character to draw by choosing a column (denoted as $c \in \mathbb{R}^d$) from the character embedding matrix $E \in \mathbb{R}^{d \times N}$, and this vector will be used at each time-step of generating (see Fig. 5(b)). The initial hidden state, pen-direction, pen-state are all set as zeros. After that, as shown in Fig. 5(b), at time-step $t$, a pen-direction $d_{t+1}$ is randomly sampled from $P_d$ in Eq. (38). Since this is a GMM model, the sampling can be efficiently implemented by first randomly choosing a component and then sampling from the corresponding Gaussian distribution. The pen-state $s_{t+1}$ is then inferred from Eq. (41) with hard-max, i.e., setting the largest element to one and remaining elements to zero.

As shown in Fig. 5(b), if the pen-state is changed to $[0, 0, 1]$ (end-of-char), the generating process should be finished. Otherwise, we should continue the drawing process by feeding $[h_t, d_{t+1}, s_{t+1}, c]$ into the GRU to generate $d_{t+2}$ and $s_{t+2}$. By repeating this process and drawing the generated lines on the paper according to the pen-states (down or up), we can obtain the automatically generated character, which should be cursive and human-readable.

## VI. Experiments on Drawing Chinese Characters

In this section, we will show the generated characters visually, and analyze the quality of the generated characters by feeding them into the discriminative RNN model to check whether they are recognizable or not. Moreover, we will also discuss properties of the character embedding matrix.

### A. Database

To train the generative RNN model, we still use the database of CASIA [43] including OLHWDB1.0 and OLHWDB1.1. There are more than two million training samples and all the characters are written cursively with frequently-used handwriting habits from different individuals. This is significantly different from the experiment in [27], where only 11,000 regular-written samples are used for training. Each character is now represented by multiple lines as shown in Eq. (23). The two hyper-parameters used in Section II for removing redundant information are $T_{\text{dist}} = 0.05 \times \max\{H, W\}$ and $T_{\cos} = 0.9$, where $H$ is the height and $W$ is the width of the character. After preprocessing, the average length of each sequence (character) is about 27. Note that the sequence used here is shorter than the sequence used for classification in Section IV. The reason for doing so is to make each line in Eq. (23) more informative and then alleviate the influence from noise strokes in the generating process.

### B. Implementation Details

Our generative RNN model is capable of drawing 3,755 different characters. The dimension for the character embedding (as shown in Section V-B) is 500. In Eqs. (25) and (26), both the low-dimensional pen-direction and pen-state are transformed to a 300-dimensional space. The dimension for the hidden state of GRU is 1000, therefore, the dimensions of the vectors in Eqs. (27)(28)(29)(30) are all 1000. The dimension for the output vector in Eq. (31) is 300, and the dropout probability applied on this output vector is 0.3. The number of mixture components in GMM of Section V-C is 30. With these configurations, the generative RNN model is trained using Adam [40] with mini-batch size 500 and initial learning rate 0.001. With Theano [44], [45] and an NVIDIA Titan-X 12G GPU, the training of our generative RNN model took about 50 hours to converge.

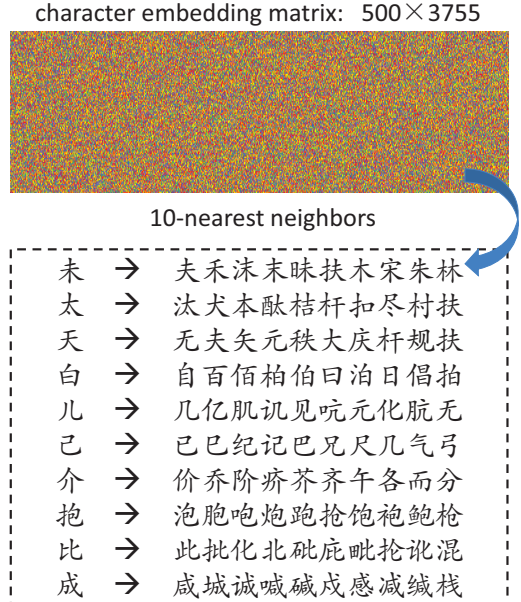

Fig. 7. The character embedding matrix and the nearest neighbors (of some representative characters) calculated from the embedding matrix.

### C. Illustration of the Training Process

To monitor the training process, Fig. 6 shows the generated characters (for the first character among the 3,755 classes) in each epoch. It is shown that in the very beginning, the model seems to be confused by so many character classes. In the first three epochs, the generated characters look like some random mixtures (combinations) of different characters, which are impossible to read. Until the 10th epoch, some initial structures can be found for this particular character. After that, with the training process continued, the generated characters become more and more clear. In the 50th epoch, all the generated characters can be easily recognized by a human with high confidence. Moreover, all the generated characters are cursive, and different handwriting styles can be found among them. This verifies the effectiveness of the training process for the generative RNN model. Another finding in the experiments is that the Adam [40] optimization algorithm works much better for our generative RNN model, compared with the traditional stochastic gradient descent (SGD) with momentum. With the Adam algorithm, our model converged within about 60 epochs.

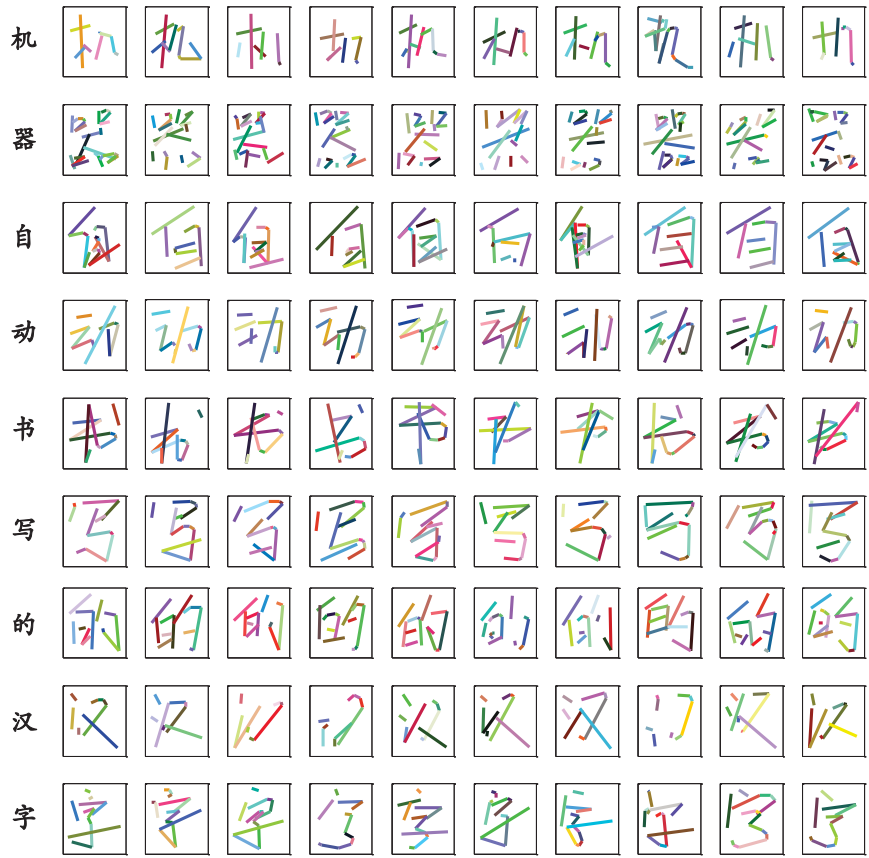

Fig. 8. Illustration of the automatically generated characters for different classes. Each row represents a particular character class. To give a better illustration, each color (randomly selected) denotes one straight line (pen-down) as shown in Eq. (23).

*D. Property of Character Embedding Matrix*

As shown in Section V-B, the generative RNN model is jointly trained with the character embedding matrix $E \in \mathbb{R}^{d \times N}$, which allows the model to generate characters according to the class indexes. We show the character embedding matrix ($500 \times 3755$) in Fig. 7. Each column in this matrix is the embedded vector for a particular class. The goal of the embedding is to indicate the RNN the identity of the character to be generated. Therefore, the characters with similar writing trajectory (or similar shape) are supposed to be close to each other in the embedded space. To verify this, we calculate the nearest neighbors of a character category according to the Euclidean distance in the embedded space. As shown in Fig. 7, the nearest neighbors of one character usually have similar shape or share similar sub-structures with this character. Note that in the training process, we did not utilize any between-class information, the objective of the model is just to maximize the generating probability conditioned on the embedding. The character-relationship is automatically learned from the handwriting similarity of each character. These results verify the effectiveness of the joint training of character embedding and the generative RNN model, which together form a model of the conditional distribution $p(x|y)$, where $x$ is the handwritten trajectory and $y$ is the character category.

*E. Illustration of Automatically Generated Characters*

With the character embedding, our RNN model can draw 3,755 different characters, by first choosing a column from the embedding matrix and then feeding it into every step of the generating process as shown in Fig. 5. To verify the ability of drawing different characters, Fig. 8 shows the automatically generated characters for nine different classes. All the generated characters are new and different from the training data. The generating/drawing is implemented randomly step-by-step, i.e., by randomly sampling the pen-direction from the GMM model as described in Section V-C, and updating the hidden states of GRU according to previous sampled handwriting trajectory as shown in Section V-B. Moreover, all the characters are automatically ended with the "end-of-char" pen-state as discussed in Section V-D, which means the RNN can automatically decide when and how to finish the writing/drawing process.

All the automatically generated characters are human-readable, and we can not even distinguish them from the real handwritten characters produced by human beings. The memory size of our generative RNN model is only 33.79MB, but it can draw as more as 3,755 different characters. This means we successfully transformed a large handwriting database (with more than two million samples) into a small RNN generator, from which we can sample infinite different characters. With the GMM modeling of pen-direction, different handwriting habits can be covered in the writing process. As shown in Fig. 8, in each row (character), there are multiple handwriting styles, e.g., regular, fluent, and cursive. These results verify not only the ability of the model in drawing recognizable Chinese characters but also the diversity of the generative model in handling different handwriting styles.

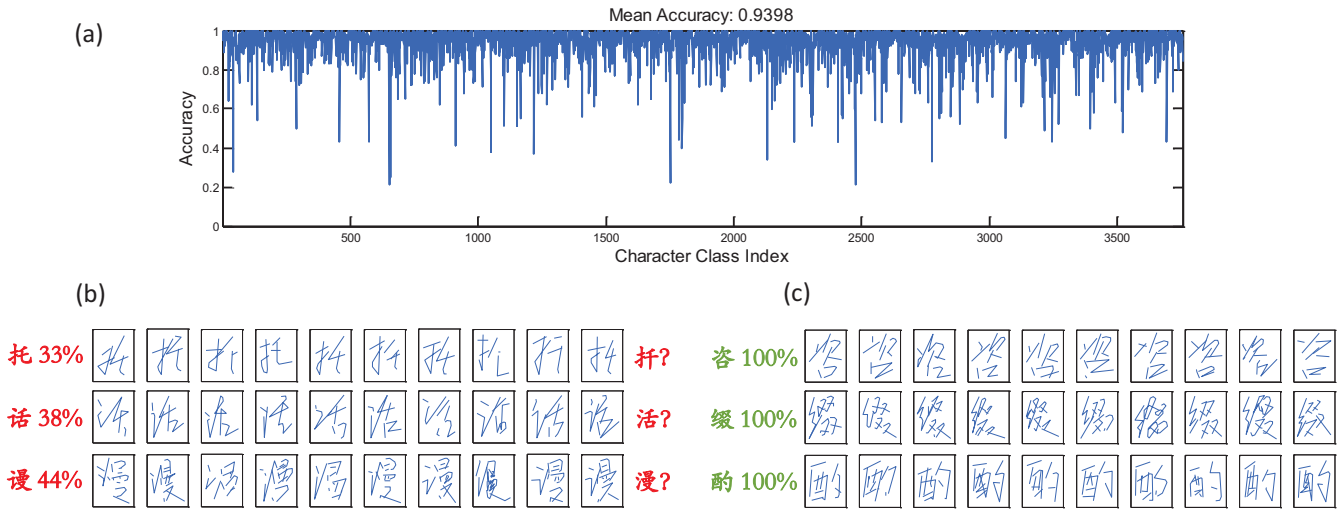

Fig. 9. (a): The classification accuracies of the automatically generated characters for the 3,755 classes. (b): The generated characters which have low recognition rates. (c): The generated characters which have perfect (100%) recognition rates.

Nevertheless, we still note that the generated characters are not 100% perfect. As shown by the last few rows in Fig. 8, there are some missing strokes in the generated characters which make them hard to read. Therefore, we must find some methods to estimate the quality of the generated characters in a quantitative manner.

*F. Quality Analysis: Recognizable or Not?*

To further analyze the quality of the generated characters, the discriminative RNN model in Section III is used to check whether the generated characters are recognizable or not. The architecture of NET4 in Table I is utilized due to its good performance in the recognition task. Both the discriminative and generative RNN models are trained with real handwritten characters from [43]. After that, for each of the 3,755-class, we randomly generate 100 characters with the generative RNN model, resulting in 375,500 test samples, which are then feed into the discriminative RNN model for evaluation. The classification accuracies of the generated characters with respect to different classes are shown in Fig. 9(a).

It is shown that for most classes, the generated characters can be automatically recognized with very high accuracy. This verifies the ability of our generative model in correctly writing thousands of different Chinese characters. In previous work of [27], the LSTM-RNN is used to generate fake and regular-written Chinese characters. Instead, in this paper, our generative model is conditioned on the character embedding, and a large real handwriting database containing different handwriting styles is used for training. Therefore, the automatically generated characters in this paper are not only cursive but also readable by both human and machine.

The average classification accuracy for all the generated characters is 93.98%, which means most of the characters are correctly classified. However, compared with the classification accuracies for real characters as shown in Table I, the recognition accuracy for the generated characters is still lower. As shown in Fig. 9(a), there are some particular classes leading to significantly low accuracies, i.e., below 50%. To check what is going on there, we show some generated characters with low recognition rate in Fig. 9(b). It is shown that the wrongly classified characters usually come from the confusable classes, i.e., the character classes which only have some subtle difference in shape with another character class. In such case, the generative RNN model was not capable of capturing these small but important details for accurate drawing of the particular character.

On the contrary, as shown in Fig. 9(c), for the character classes which do not have any confusion with other classes, the generated characters can be easily classified with 100% accuracy. Therefore, to further improve the quality of the generated characters, we should pay more attention to the similar/confusing character classes. One solution is to modify the loss function in order to emphasize the training on the confusing class pairs as suggested by [47]. Another strategy is integrating the attention mechanism [48], [49] and the memory mechanism [50], [51] with the generative RNN model to allow the model to dynamically memorize and focus on the critical region of a particular character during the writing process. These are future directions for further improving the quality of the generative RNN model.

## VII. Conclusion and Future Work

This paper investigates two closely-coupled tasks: automatically reading and writing. Specifically, the recurrent neural network (RNN) is used as both discriminative and generative models for recognizing and drawing cursive handwritten

Chinese characters. In the discriminative model, the deep stacked bidirectional RNN model is integrated with both the LSTM and the GRU for recognition. Compared with previous convolutional neural network (CNN) based approaches which require some image-like representations, our method is fully end-to-end by directly dealing with the raw sequential data. Due to the straightforward utilization of the spatial and temporal information, our discriminative RNN model achieved new state-of-the-art performance on the ICDAR-2013 competition database. High character recognition accuracy is essential for text recognition [52], [53], hence, the discriminative RNN model can be hopefully combined with the CTC [31], [18] for segmentation-free handwritten Chinese text recognition [54]. Moreover, another potential direction is combining the powerful image processing ability of CNNs and the sequence processing ability of RNNs to look for further accuracy improvement on character recognition.

Besides recognition, this paper also considers automatic drawing of real and cursive Chinese characters. A conditional generative RNN model is jointly trained with the character embedding, which allow the model to correctly write more than thousands of different characters. The Gaussian mixture model (GMM) is used for modeling the pen-direction which guarantee the diversity of the model in generating different handwriting styles. The generative RNN model can automatically decide when and how to finish the drawing process with the modeling of three discrete pen-states. It is shown that the generated characters are not only human-readable but also recognizable by the discriminative RNN model with high accuracies. Other than drawing characters, an interesting future direction is to utilize the proposed method as building blocks for the synthesis of cursive handwritten Chinese texts. Moreover, in this paper, the generative model is conditioned on the character embedding. Another important future extension is to condition the generative RNN model on a static image (combined with convolution) and then automatically recover (or generate) the dynamic handwriting trajectory (order) from the static image [55], [56], which is a hard problem and has great values in practical applications.

The relationship between the discriminative and generative models is also an important future research topic. Since the generative model is capable of producing real handwritten characters (with labels), a straightforward attempt is to make it as a data augmentation strategy for the supervised training of the discriminative model. On the opposite direction, the discriminative model can be used as some regularization [57] to improve the quality of the generative model. Moreover, the generative model can also cooperate with the discriminative model in an adversarial manner [23], [58]. Taking all these together, an attractive and important future work is the simultaneously training of the discriminative and generative models in an unified multi-task framework.

ACKNOWLEDGMENTS

The authors thank the developers of Theano [44], [45] for providing such a good deep learning platform.